%% file: main.tex
\documentclass[10pt,twocolumn,letterpaper]{article}

\usepackage[pagenumbers]{cvpr}              

\usepackage{multirow}
\usepackage{array}
\usepackage{siunitx}
\usepackage{tabularray}
\input{preamble}
\definecolor{cvprblue}{rgb}{0.21,0.49,0.74}
\usepackage[pagebackref,breaklinks,colorlinks,allcolors=cvprblue]{hyperref}

\def\confName{CVPR}

\title{UniLoc: Towards Universal Place Recognition Using Any Single Modality}

\author{
    \small
    \begin{tabular}{c c c c c c c}                                            
        Yan Xia$^{*1,2 \dagger}$ &  
        Zhendong Li$^{*1}$ &  
        Yun-Jin Li$^1$ &  
        Letian Shi$^1$ &  
        Hu Cao$^1$ &  
        João F. Henriques$^3$ &  
        Daniel Cremers$^{1,2}$ \\                                        
        \multicolumn{7}{c}{\shortstack{$^1$Technical University of Munich \quad $^2$Munich Center for Machine Learning (MCML) \\ 
        $^3$ Visual Geometry Group, University of Oxford}} \\                                                
        \multicolumn{7}{c}{\tt\small \{yan.xia, zhendong.li,yunjin.li,letian.shi,cao.hu,cremers\}@tum.de, joao@robots.ox.ac.uk} 
    \end{tabular}
    }
\begin{document}
\twocolumn[{
\renewcommand\twocolumn[1][]{#1}%
\renewcommand{\thefigure}{\arabic{figure}}
\maketitle
\begin{center}
\captionsetup{type=figure}
\includegraphics[width=1.0\linewidth]{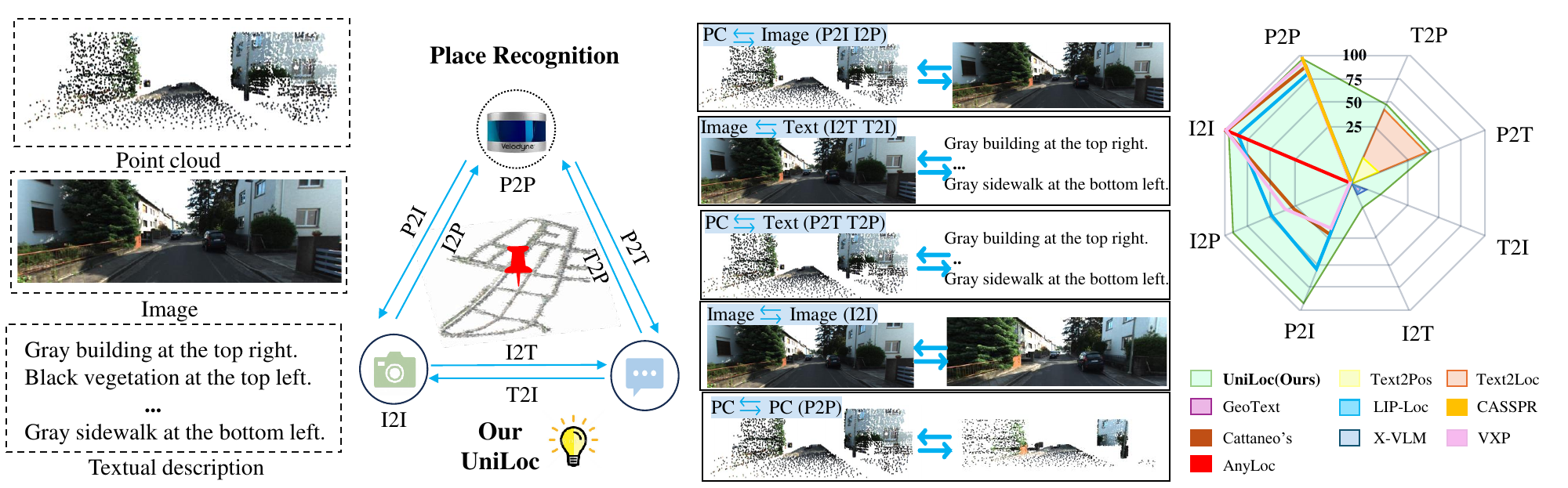}
\captionof{figure}{\textit{(Left)} We present UniLoc, a solution designed for city-scale place recognition using any one of three modalities: text, image, or point cloud.
\textit{(Right)} Localization performance at top-1 recall on the KITTI-360~\cite{liao2022kitti} test set. The proposed UniLoc achieves state-of-the-art performance across all six cross-modality place recognition: image-to-point cloud (I2P), point cloud-to-image (P2I), text-to-point cloud (T2P), point cloud-to-text (P2T), image-to-text (I2T), and text-to-image (T2P). Notably, UniLoc surpasses existing SOTA cross-modal methods by a large margin while achieving competitive performance in uni-modal place recognition.}
\label{fig:cover}
\end{center}%

}]
\renewcommand{\thesection}{\arabic{section}}
\renewcommand{\thefigure}{\arabic{figure}}
\renewcommand{\thetable}{\arabic{table}}
\let\oldthefootnote\thefootnote
\renewcommand{\thefootnote}{\fnsymbol{footnote}} 
\footnotetext[2]{Corresponding author. * Equal contribution.} 
\let\thefootnote\oldthefootnote
\input{sec/0_abstract}    
\input{sec/1_intro}

\input{sec/2_formatting}

\input{sec/3_Problem_Statement}
\input{sec/5_Methodology}

\input{sec/4_Dataset_Generation}

\input{sec/6_Experiments}
\input{sec/7_Conclusion}


{
    \small
    \bibliographystyle{ieeenat_fullname}
    \bibliography{main}
}

\setcounter{section}{0} %
\renewcommand{\thesection}{\Alph{section}} %
\input{sec/X_suppl}

\end{document}

%% file: preamble.tex
\usepackage[utf8]{inputenc} 
\usepackage[T1]{fontenc}    
\usepackage{url}            
\usepackage{booktabs}       
\usepackage{amsfonts}       
\usepackage{fontawesome}
\usepackage{nicefrac}       
\usepackage{microtype}      
\usepackage{xcolor}  
\usepackage{multirow}
\usepackage{longtable}
\usepackage{supertabular}
\usepackage{makecell}
\usepackage{soul}
\usepackage{xcolor}
\usepackage{amsmath}
\usepackage{amssymb}
\usepackage{bbm}
\usepackage{graphicx}
\usepackage{subcaption}
\usepackage{wrapfig}
\usepackage{colortbl}
\usepackage{adjustbox}
\usepackage{siunitx}
\renewcommand\thesection{\Alph{section}}

\renewcommand{\thefigure}{\Roman{figure}}
\renewcommand{\thetable}{\Roman{table}}

%% file: sec/0_abstract.tex
\begin{abstract}
To date, most place recognition methods focus on single-modality retrieval. While they perform well in specific environments, cross-modal methods offer greater flexibility by allowing seamless switching between map and query sources.
It also promises to reduce computation requirements by having a unified model, and achieving greater sample efficiency by sharing parameters.
In this work, we develop a universal solution to place recognition, UniLoc, that works with any single query modality (natural language, image, or point cloud).
UniLoc leverages recent advances in large-scale contrastive learning, and learns by matching hierarchically at two levels: instance-level matching and scene-level matching. 
Specifically, we propose a novel Self-Attention based Pooling (SAP) module to evaluate the importance of instance descriptors when aggregated into a place-level descriptor.
Experiments on the KITTI-360 dataset demonstrate the benefits of cross-modality for place recognition, achieving superior performance in cross-modal settings and competitive results also for uni-modal scenarios. Our project page is publicly available at 
\url{https://yan-xia.github.io/projects/UniLoc/}. 
\end{abstract}

%% file: sec/1_intro.tex
\section{Introduction}
\label{sec:intro}

Place Recognition (PR) is essential for autonomous vehicles and robots, enabling them to self-localize accurately in complex, large-scale environments, such as urban areas or parking structures where buildings or tunnels can block Navigation Satellite System (GNSS) signals. The goal of place recognition is to identify the closest matching location in a pre-built database from a given query that describes the current scene. Given the range of sensors onboard robots, queries can take different forms, including natural language descriptions, RGB images, or LiDAR point clouds.

Usually, the implementation of place recognition is expressed as a retrieval task. Over the past decade, image retrieval-based solutions~\cite{arandjelovic2016netvlad, noh2017large, ali2023mixvpr, keetha2023anyloc} have shown promising results. However, these methods often struggle with significant variations in lighting, weather, and seasonal appearances. As a possible remedy, point cloud based place recognition~\cite{angelina2018pointnetvlad, xia2021soe, komorowski2021minkloc3d, Xia_2023_ICCV} becomes an attractive research topic since 3D point clouds acquired from LiDAR provide highly accurate, detailed, and illumination-invariant spatial information~\cite{xia2021vpc}. Yet, despite its robustness, relying on only a single data source like LiDAR poses limitations, particularly if sensors malfunction or if the sensor setup varies across environments~\cite{li2024vxp}.

To address these limitations, cross-modal methods~\cite{2d3dmatch,li2024vxp, kolmet2022text2pos, xia2024text2loc,zeng2022multi,chu2025towards} have emerged, offering flexibility by enabling place recognition across different query and map data sources. For example, camera-to-LiDAR approaches~\cite{2d3dmatch,li2024vxp} support querying a LiDAR database with 2D images, which is beneficial in situations where processing large point clouds is impractical or where LiDAR data is unavailable or compromised. Such approaches reduce computational load and ensure reliable localization even with limited sensor availability. 
This cross-modal capability is also highly relevant to real-world navigation tasks, such as delivery services, where couriers may rely on verbal directions from recipients to find precise drop-off points. More broadly, the “last mile problem”—navigating to an unfamiliar final destination—highlights the need for flexible place recognition methods that can incorporate natural language input~\cite{kolmet2022text2pos,xia2024text2loc,zeng2022multi,chu2025towards}. Despite this demand, there remains no universal place recognition solution that seamlessly integrates text, images, and point clouds. 

In this work, we address the question: \textbf{how can we design a universal place recognition solution that works regardless of the query modality?}
This entails generating shared place representations from a general model by leveraging any single-modal data, including text, image, and point clouds. 
Unlike ImageBind~\cite{girdhar2023imagebind}, which is more focused on instance-level retrieval and lacks geographic information, the main challenge in ours lies in mapping the \textit{place-level} descriptors from various modalities into a unified space, enabling consistent and accurate retrieval.

As highlighted by VXP~\cite{li2024vxp}, local correspondences are crucial for accurate global localization. Inspired by this, we decompose the place recognition problem into two stages: instance-level matching and scene-level matching. In instance-level matching, we aim to map pairs of modalities—(image, text) and (image, point cloud)—into a shared learning space. The reason is that ImageBind has shown aligning each modality’s embedding to image embeddings results in emergent cross-modal alignment.
We focus on fusing multiple modalities and follow recent works that perform learning over large pre-trained models, such as BLIP-2~\cite{li2023blip}.
In scene-level matching, we find that different instance descriptors should tactically contribute unevenly when aggregated into a place-level descriptor. To achieve this, we develop a Self-Attention based Pooling (SAP) module instead of the max-pooling layers widely used in previous global descriptors~\cite{angelina2018pointnetvlad, xia2021soe}. 
These operations enable one-stage training to map the place-level descriptors from different modalities into a shared embedding space.

To summarize, the main contributions of this work are:

\begin{itemize}
    \item To the best of our knowledge, we are the first to propose a universal place recognition network for large-scale outdoor environments that can handle any single modality, including natural language, images, or point clouds.
    \item We propose a novel method UniLoc, leveraging recent advances in large-scale contrastive learning and following hierarchically matching stages. 
    \item We propose a novel self-attention based pooling method that directs the model to focus on the more discriminative instance descriptors during the scene-level matching.
    \item We conduct extensive experiments on the KITTI-360 data set~\cite{liao2022kitti} and show that the proposed UniLoc greatly improves over the state-of-the-art cross-modal methods while achieving competitive performance compared with uni-modal methods.
\end{itemize}

%% file: sec/2_formatting.tex
\begin{figure*}
    \centering
    \includegraphics[width=1\linewidth]{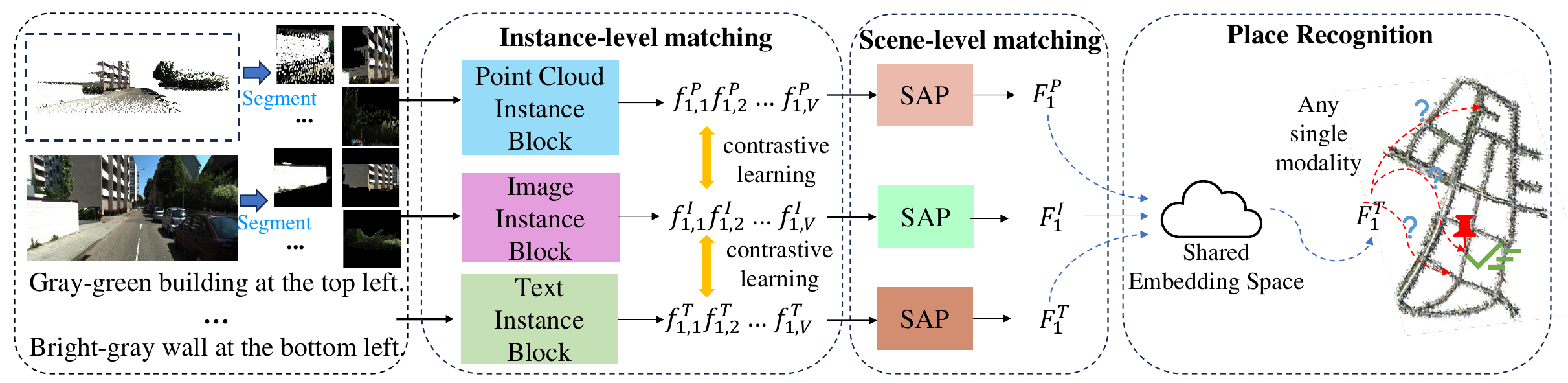}
    \caption{Overview of the proposed pipeline, consisting of instance-level (sec. \ref{instance block}) and scene-level (sec. \ref{scene model}) matching stages.}
    \label{fig:pipline}
\end{figure*}

\section{Related Work}
\label{sec: related work}
We will focus on the related tasks of visual place recognition, point cloud based place recognition, and cross-modal place recognition between two different modalities.

\noindent
\textbf{Visual place recognition.} Given a query image or image sequence, visual place recognition (VPR) aims to retrieve the closest match to this query on the geo-tagged reference map. The effectiveness of modern VPR methods is primarily due to training in large-scale datasets. NetVLAD~\cite{arandjelovic2016netvlad} achieved excellent performance by employing weakly-supervised contrastive learning with the Pitts250k dataset~\cite{torii2013visual}.  Subsequently,  DeLF~\cite{noh2017large} and DeLG~\cite{cao2020unifying} explore the large-scale image retrieval tasks supported by the Google-Landmark V1 (1 million images) and V2 (5 million images) datasets~\cite{weyand2020google}. Unlike them, CosPlace~\cite{berton2022rethinking} explores retrieving as a classification task. MixVPR~\cite{ali2023mixvpr} proposes extracting features using an MLP-based feature mixer. More recently, AnyLoc~\cite{keetha2023anyloc} is the first VPR solution that exhibits anywhere, anytime, and anyview capabilities. However, the performance of VPR often declines under significant variations in illumination and appearance due to changes in weather and seasons. 

\noindent
\textbf{Point cloud based place recognition.}
Compared with images, point cloud based place recognition provides distinct advantages, remaining robust against variations in lighting, weather, and seasons.
PointNetVlad~\cite{angelina2018pointnetvlad} was a pioneering network using end-to-end learning for 3D place recognition. Building on this, SOE-Net~\cite{xia2021soe} introduces the PointOE module, adding orientation encoding to PointNet for point-wise local descriptors. Additionally, several methods~\cite{zhou2021ndt, deng2018ppfnet, fan2022svt, zhang2022rank, ma2022overlaptransformer, barros2022attdlnet, ma2023cvtnet} incorporate transformer networks with stacked self-attention blocks to capture long-range contextual features. In contrast, Minkloc3D~\cite{komorowski2021minkloc3d} uses a voxel-based approach with a Feature Pyramid Network (FPN)\cite{lin2017feature} and GeM pooling\cite{radenovic2018fine} to create a compact global descriptor, though voxelization can lose points during quantization. To address this, CASSPR~\cite{Xia_2023_ICCV} introduces a dual-branch hierarchical cross-attention transformer, blending the benefits of voxel- and point-based methods.

\noindent
\textbf{Cross-modal place recognition.} With breakthroughs in large language models (LLM), vision and language localization have become an intense research focus. Language-guided localization requires an agent to follow natural language instructions when localizing in a specific environment. X-VLM~\cite{zeng2022multi} achieves Image-Text retrieval by aligning text descriptions with the
corresponding visual concepts in images. GeoText~\cite{chu2025towards} is the first work to explore the language-guided drone navigation task by text-to-image and image-to-text retrieval. For Image-LiDAR place recognition, Cattaneo \etal \cite{2d3dmatch} first establish a shared global feature space between 2D images and 3D point clouds using a teacher-student model.
Recent VXP~\cite{li2024vxp} improves the retrieval performance by enforcing local similarities in a self-supervised manner.
In Text-LiDAR point cloud localization, Text2Pos~\cite{kolmet2022text2pos} pioneered large-scale outdoor scene localization by first identifying coarse locations and then refining pose estimates. Building on this, Wang \etal~\cite{wang2023text} introduced a Transformer-based approach to improve representation discriminability for both point clouds and text queries. Text2Loc~\cite{xia2024text2loc} further enhances localization accuracy and efficiency by adopting contrastive learning and a lighter, faster strategy that eliminates the need for a text-instance matcher. In this work, we aim to address the place recognition problem in a unified solution that leverages any single modality independently.

%% file: sec/3_Problem_Statement.tex
\section{Problem Statement}

We begin by defining the large-scale  point cloud reference map ${ PC_\textrm{ref} = {\left \{S_{i}: i = 1,..., M \right \}}}$ , which is a collection of submaps $S_i$. Each 3D submap ${ S_i = {\left \{p_{i,j} \mid j = 1, \ldots, n \right\}} }$ contains a set of 3D object instances $p_{i,j}$.
Let ${ I_\textrm{ref} = {\left \{I_i \mid i = 1, \ldots, M \right\}}}$ represent an image sequence, where each image $I_i$ corresponds to the same location as the submap $S_i$ in $PC_\textrm{ref}$. Each 2D image ${ I_i = { \left \{o_{i,j} \mid j = 1, \ldots, n \right \}} }$ contains a set of 2D object instances $o_{i,j}$. 
In our work, an image $I_i$ has the same number of object instances as $S_i$.
Besides, let ${ T_\textrm{ref} = {\left \{T_i \mid i = 1, \ldots, M \right\}}}$
be text descriptions. Each  $T_i$ includes a set of hints $\{\Vec{h}_k\}_{k=1}^{h}$, with each hint describing the spatial position of an object instance in $I_i$ space.

We approach this task by decomposing it into two stages: instance-level matching and scene-level matching. The instance-level matching stage aims to train a function $F$ that aligns individual object instances across different modalities. Specifically, we align each 3D object instance $p_{i,j}$ with its corresponding 2D representation $o_{i,j}$ and also align the 2D object instance $o_{i,j}$ with its corresponding text hint $\vec{h}_k$. The function $F$ encodes these 3D, 2D, and textual representations into a unified embedding space. In this embedding space, matching instances across modalities are brought closer together, while non-matching instances are pushed apart.
In the context of scene-level matching, we leverage the feature extraction capabilities of the pre-trained instance-level model. The goal is to train a function \( M \) that effectively aggregates the instance-level descriptors into a unified place-level descriptor. 

%% file: sec/5_Methodology.tex
\section{UniLoc}
To the best of our knowledge, our approach, UniLoc, is the first place recognition solution that exhibits any single modality (text, image, or point cloud). 
Fig. \ref{fig:pipline} shows the pipeline of our UniLoc.
Given a triplet consisting of text descriptions, a single image, and the corresponding 3D submap, we first align the image with the point cloud and the image with the text at the instance level, as detailed in Section \ref{instance block}.
Next, we aggregate the instance-level descriptors into a place-level representation, as discussed in Section \ref{scene model}. The training strategy and loss function are detailed in Section \ref{loss function}.

\subsection{Instance-level matching}
\label{instance block}
\begin{figure}
    \centering
\includegraphics[width=1\linewidth]{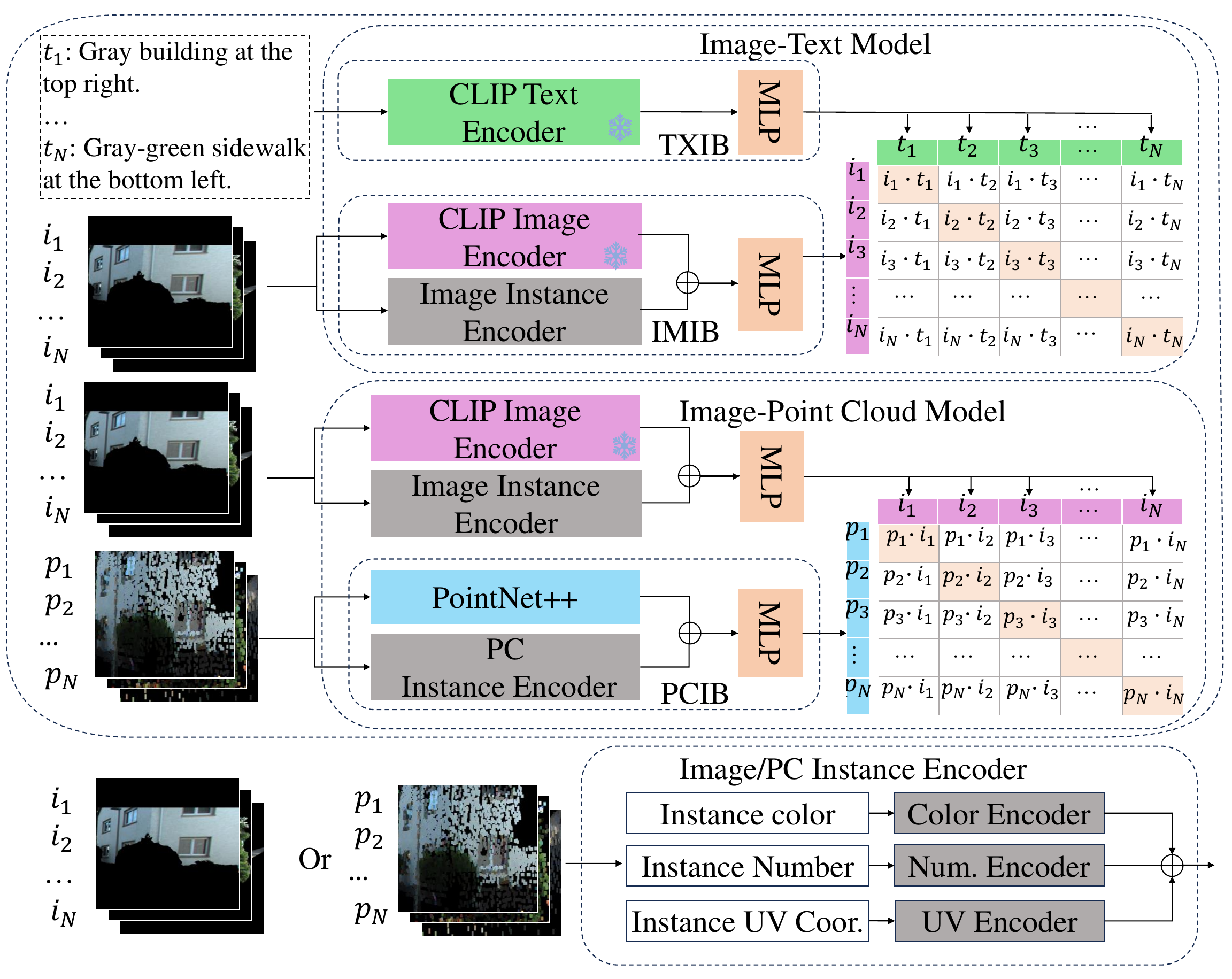}
    \caption{\textit{(Top)} The architecture of instance-level matching. 
    It consists of three instance-level feature extraction blocks: Text Instance Block (TXIB), Image Instance Block (IMIB), and Point Cloud Instance Block (PCIB). We train an Image-Text and an Image-Point cloud model to align image-text instances and image-point cloud instances, respectively. \textit{(Bottom)} The architecture of the image and point cloud instance encoders. Note that the pre-trained CLIP image and text encoders are frozen during training.}
    \label{fig:Instance encoder}
\end{figure}

Considering the stability of training, UniLoc adopts an image-centric approach, where other modalities are embedded into a shared embedding space that aligns closely with the image representation. To achieve this, we employ two instance-level models during the pretraining stage, both centered around image instances: the Image-Text Model and the Image-Point Cloud Model 
as illustrated in Fig. \ref{fig:Instance encoder}. Below, we provide a detailed explanation of the instance branches that constitute these two models.

\noindent
\textbf{Text branch.} 
We utilize a pre-trained language encoder in CLIP~\cite{radford2021learning} to extract nuanced features from textual descriptions of object instances. We then design a three-layer MLP to transform the features into a text space \( \mathbf{f}_i^{T} \in \mathbb{R}^{1 \times D} \), where \( D \) indicates the feature dimension.

\noindent
\textbf{Image branch.} 
We use a frozen pre-trained image encoder from CLIP to extract semantic information for each instance, followed by a three-layer MLP to project these features into the image space.
Inspired by Text2Loc~\cite{xia2024text2loc}, to capture additional attributes, we utilize an image instance encoder consisting of three components: the Color Encoder, which takes the average RGB values of non-zero regions; the Number Encoder, which uses the normalized count of non-zero pixels; and the UV Encoder, which encodes the average UV coordinates within the original image. Each encoder is a three-layer MLP.
Finally, all features are concatenated and passed through another three-layer MLP to produce the image instance descriptor $\mathbf{f}_i^{I} \in \mathbb{R}^{1 \times D}$,
where \( D \) indicates the embedding dimension.


\noindent
\textbf{Point cloud branch.}
Following~\cite{xia2024text2loc}, we first use PointNet++\cite{qi2017pointnet} (although other point networks could also be used) to extract the semantic features from 3D instances.
We additionally obtain a color embedding by encoding the average RGB values of the points using a color encoder. 
We also introduce a number encoder to provide potential class-specific prior information by explicitly encoding the number of points.
Unlike Text2Loc~\cite{xia2024text2loc}, we design a novel positional embedding by encoding the UV coordinates of the points (details on the UV coordinate calculation are provided in Section \ref{data generation}), similar to what we used in the image branch.
The color, positional, and number encoders are all 3-layer multi-layer perceptrons (MLPs) with output dimensions matching the semantic point embedding dimension.
We concatenate the semantic, color, positional, and number embeddings, and then process them with a projection layer, another 3-layer MLP, to obtain the point cloud instance descriptor $\mathbf{f}_i^{P} \in \mathbb{R}^{1 \times D}$. 


\subsection{Scene-level matching}
\label{scene model}
\begin{figure*}
    \centering
    \includegraphics[width=0.8\linewidth]{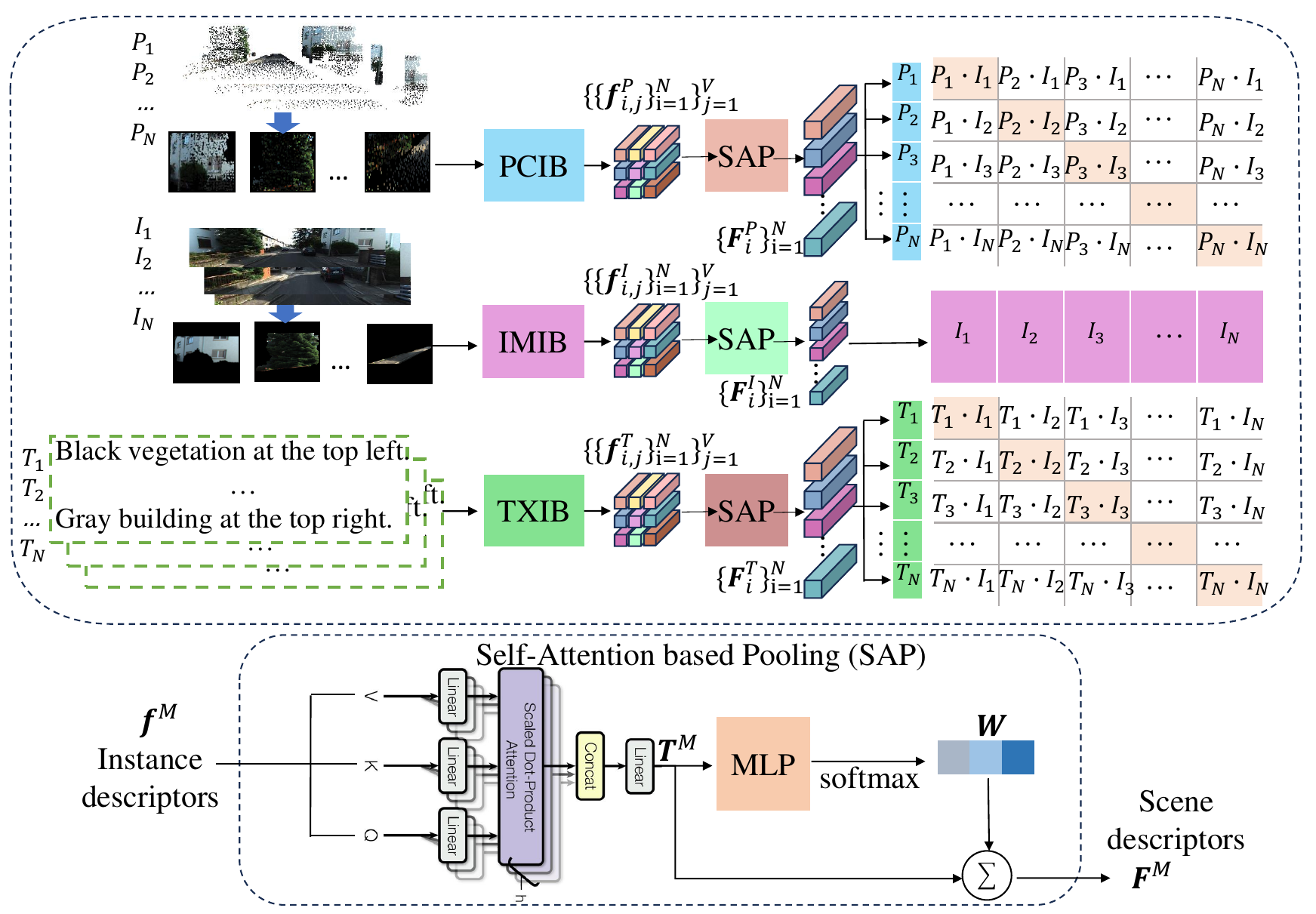}
    \caption{\textit{(Top)} The proposed fine matching architecture of UniLoc. It consists of triple parallel feature extraction branches: Point cloud, image, and text. \textit{(Bottom)} The architecture of the proposed Self-Attention based Pooling (SAP) module.}
    \label{fig:uni loc}
\end{figure*}

Following the instance-level matching, we aim to create a shared embedding
space between textual descriptions, images, and LiDAR point clouds via aggregating the single-modality instance descriptors to a place-level descriptor. The detailed structures are shown in Fig. \ref{fig:uni loc} (Top).
We first load the pre-trained weights from the instance-level matching phase to extract each instance descriptor. Specifically, we use the Image-Text Model weights for the current image and text branches, while the point cloud branch uses the pre-trained weights from the Image-Point Cloud Model. We then introduce a Self-Attention based Pooling (SAP) module to aggregate these instance descriptors into a robust and discriminative scene descriptor.

\noindent \textbf{Self-Attention based Pooling (SAP).} 
Inspired by SOE-Net~\cite{xia2021soe}, we find that
different instance descriptors should have varying importance when aggregating a global scene descriptor. For example, a building can be much less distinctive of a specific scene than a traffic light. We thus propose a self-attention based pooling mechanism intended to replace the previous widely used max-pooling layers in ~\cite{xia2021soe}.
Multi-Head Self-Attention (MHSA) and FFN sublayers \cite{xia2024text2loc} \cite{vaswani2017attention} take the instance descriptor \(\mathbf{f}^{M}\) of any modality \(M\) as input and computes the attention feature $\mathbf{T}^{M} \in \mathbb{R}^{V \times D}$:
\begin{align}
    \tilde{\mathbf{f}}^{M} &= \mathbf{Q} + \text{MSHA}(\mathbf{Q}, \mathbf{K}, \mathbf{V}), \\
    \mathbf{T}^{M} &= \text{Transformer}(\mathbf{Q}, \mathbf{K}, \mathbf{V}) = \left[ \tilde{\mathbf{f}}^{M} + \text{FFN}(\tilde{\mathbf{f}}^{M}) \right]
\end{align}
where $\mathbf{Q}$, $\mathbf{K}$, and $\mathbf{V}$ are the query, key, and value matrices respectively, and for simplicity we use the instance descriptor $\mathbf{f}^{M}$ to fill their roles: $\mathbf{Q} = \mathbf{K} = \mathbf{V} = \mathbf{f}^{M} \in \mathbb{R}^{V \times D}$, where $V$ is the number of instances, and $D$ is the embedding dimension.\\
We then use a 3-layer MLP and softmax operators to derive attention weights  $\mathbf{W} \in \mathbb{R}^{V \times 1}$ by taking $\mathbf{T^{M}}$ as the input:
\begin{equation}
\mathbf{W} = \text{Softmax}(\text{MLP}(\mathbf{T^{M}})),
\end{equation}
The scene descriptor \(\mathbf{F}^{M}\) is finally obtained by aggregating \(\mathbf{T}^{M}\) using the attention weights:
\begin{equation}
\mathbf{F}^{M} = \sum_{i=1}^{V} \mathbf{W}_i \mathbf{T}_i^{M},
\end{equation}
where $\mathbf{W}_i$ is the attention wights for the $i$-th instance, $\mathbf{T}_i^{M} \in \mathbb{R}^{1\times D}$ is the corresponding attention feature. SAP guides the more important instance descriptors to contribute more to the
representation of the target scene descriptor.




\subsection{Loss function}
\label{loss function}
\textbf{Instance-level matching.}
\noindent We use instance pairs \((I_{ins}, T_{ins}, P_{ins})\), where \(I_{ins}\) and \(P_{ins}\) represent the 2D and 3D instances, respectively, and \(T_{ins}\) represents the textural description of the instances. The contrastive loss \cite{radford2021learning} between \((I_{ins}, T_{ins})\) can be calculated using the following formula:
\begin{equation}\footnotesize
    l(i, I_{ins}, T_{ins}) = - \log \frac{\exp(\mathbf{f}_i^I \cdot \mathbf{f}_i^T / \tau)}{\sum\limits_{j \in N} \exp(\mathbf{f}_i^I \cdot \mathbf{f}_j^T / \tau)} 
    \\ - \log \frac{\exp(\mathbf{f}_i^T \cdot \mathbf{f}_i^I / \tau)}{\sum\limits_{j \in N} \exp(\mathbf{f}_i^T \cdot \mathbf{f}_j^I / \tau)}
\end{equation}
\noindent where \( \mathbf{f}_i^{I} \) represents the image instance descriptors, and \( \mathbf{f}_i^{T} \) denotes the descriptors of the corresponding text. \( N \) is the batch size, and \( \tau \) is a temperature parameter. The batch loss during training is obtained by averaging each contrastive loss term:
\begin{equation}
    L(I_{ins}, T_{ins}) = \frac{1}{N} \left[ \sum_{i \in N} l_I(i, I, T) \right]
\end{equation}
Similarly, \( L(I_{ins}, P_{ins}) \) can be derived in the same way.

\noindent
\textbf{Scene-level matching.}
As a modality that is inexpensive to collect, images can easily align with various sensory modalities. Specifically, for our task, a large amount of image data, such as Google Maps imagery, is available for aligning with point cloud and textual data. Given these reasons, similar to \cite{girdhar2023imagebind}, we adopt an image-centric approach where other modalities align through image embeddings, effectively creating a shared embedding space.  

\noindent We use pairs \((I, T)\), where \(I\) represents images and \(T\) represents the corresponding textual descriptions. The contrastive loss between images and text can be calculated using the following formula:
\begin{equation}\footnotesize
    l(i, I, T) = - \log \frac{\exp(F_i^I \cdot F_i^T / \tau)}{\sum\limits_{j \in N} \exp(F_i^I \cdot F_j^T / \tau)} 
    - \log \frac{\exp(F_i^T \cdot F_i^I / \tau)}{\sum\limits_{j \in N} \exp(F_i^T \cdot F_j^I / \tau)}
\end{equation}

\noindent where \( F_i^{I} \) represents the image descriptor, and \( F_i^{T} \) denotes the the corresponding text descriptor. \( N \) is the batch size. 
The batch loss during training is obtained by averaging each contrastive loss term:
\begin{equation}
    \label{final loss}
    L(I, T) = \frac{1}{N} \left[ \sum_{i \in N} l(i, I, T) \right]
\end{equation}
\( L_S(I, P) \) can be derived in the same way.

\noindent Similar to \cite{xue2023ulip}, we combine the contrastive loss between images and point clouds and between images and texts through a weighted approach to achieve alignment and proximity among the three modalities in the embedding space:
\begin{equation}
    L_{final} = \alpha L(I,T) + (1-\alpha)L(I,P) 
\end{equation}
where \(L{(I,T)} \) and \( L{(I,P)} \) represent the contrastive losses of image-text and image-point cloud pairs, $\alpha$ is the hyperparameters to balance their relationship.   


%% file: sec/4_Dataset_Generation.tex
\section{Experimental Dataset}
\label{data generation}
Following \cite{xia2024text2loc}, we use nine sequences from the KITTI-360 dataset \cite{liao2022kitti} as experimental data, with five sequences (00, 02, 04, 06, 07) for training, one (10) for validation, and three (03, 05, 09) for testing. The KITTI-360 dataset spans a driving distance of 80 km with a total coverage area of 15.51 km². Additionally, KITTI-360 labels static objects (e.g., poles, buildings, fences, garages), which makes it particularly suitable for our task.
We establish a one-to-one correspondence among the point clouds, text descriptions, and images. To ensure sufficient textual descriptions for image-text and point cloud-text retrieval in both training and evaluation, we filtered out data with fewer than six instances following ~\cite{xia2024text2loc}. In total, we obtain 19,866 triplets of text, image, and point cloud data for training, 1,781 triplets for validation, and 7,791 triplets for testing. More details are in Appendix A.

\noindent
\textbf{Images.} In the training stage,  we segment each instance in the raw image using ground-truth semantic labels. To align textual descriptions accurately with instances, we identify connected regions in the instance mask and select the largest as the instance mask. Instances with fewer than 50 valid pixels were deemed unobservable and excluded. The segmented regions are then resized to 224×224, maintaining the aspect ratio and padding with zeros as needed. We use an off-the-shelf segmentation network PSPNet \cite{zhao2017pyramid} for image segmentation during inference, in the text-image modalities experiment.

\noindent
\textbf{Point clouds.} 
We initially sample point clouds within a \SI{40}{m} radius centered on the location of each image.
Using the camera's pose matrix, we transform these points into the camera's coordinate system and project them onto the image plane using intrinsic parameters to obtain UV coordinates. We then extract the point cloud segments corresponding to each image instance by utilizing the image bounding boxes and the semantic IDs of the point cloud points.
Instances too small to have valid point cloud projections are excluded from both datasets.

\noindent
\textbf{Text.} 
We describe each image and 3D submap by combining text descriptions of its instances. Each instance's text hint includes the color, category, and position information. 
The average RGB values of each instance in the image are used for color description. 
We assign the category description based on the ground-truth semantic label. 
For the position, we define six regions based on the average UV coordinates of each instance. We divide the V-axis into ``top'' and ``bottom'' and the U-axis into ``left'' (0 to 0.4 times the image width), ``center'' (0.4 to 0.6), and ``right'' (0.6 to 1).  



%% file: sec/6_Experiments.tex
\section{Experiments}
\subsection{Evaluation criteria}
Following \cite{xia2024text2loc}, we evaluate the place recognition capability of our UniLoc across different modalities or within the same modality using Retrieval Recall at Top $k$ $(k \in {1, 3, 5})$.
For a fair comparison in Image-LiDAR, Image-Image, and LiDAR-LiDAR place recognition, we regard the retrieved scan as a correct match if the distance is within \SI{20}{m}, following \cite{li2024vxp,keetha2023anyloc,angelina2018pointnetvlad}. For Text-Image and Text-LiDAR place recognition, retrieving only the image/point cloud at the exact location is considered successful.

\subsection{Results}
\subsubsection{Text-Image place recognition}
\begin{figure}[!htbp]
    \centering
    \includegraphics[width=\linewidth]{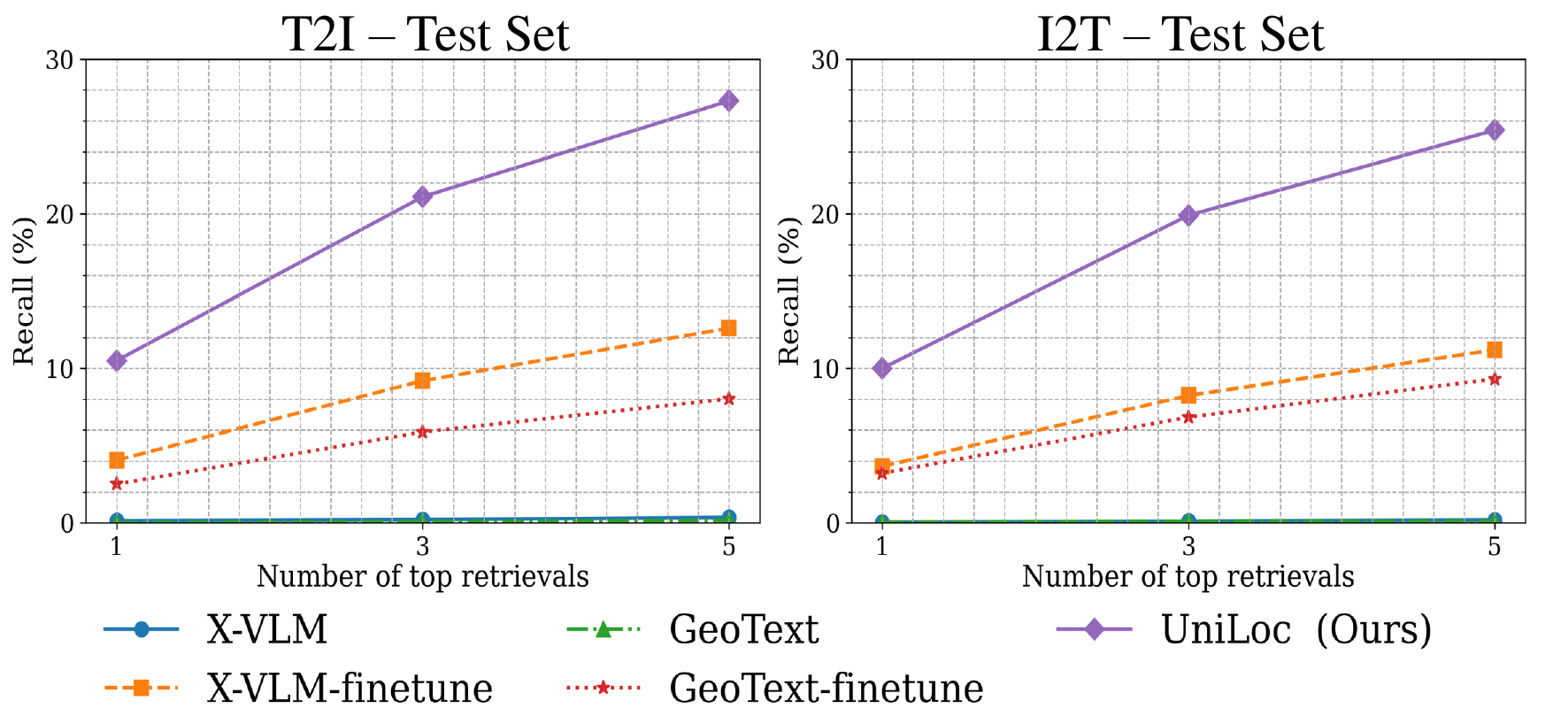}
    \caption{Performance comparison for Text-Image place recognition on the KITTI-360 dataset. "X-VLM/GeoText-finetune" indicates that we finetune the model on the KITTI-360 dataset.}
    \label{fig:image-text}
\end{figure}

We compare our UniLoc with state-of-the-art image-text retrieval methods X-VLM~\cite{zeng2022multi} and GeoText~\cite{chu2025towards} on the test set. 
Fig. \ref{fig:image-text} shows the top 1/3/5 recall of each method. 
UniLoc achieves the best performance across all recall levels.
For Text-to-Image retrieval, UniLoc outperforms the recall achieved by the fine-tuned X-VLM  by a wide margin of 6.4\% at top-1.
Furthermore, for the Image-to-Text retrieval, UniLoc surpasses X-VLM with increases of 6.3\%, 11.7\%, and 14.2\% on the test set at top-1/3/5, respectively. These consistent improvements demonstrate our UniLoc marks a clear advancement in the Text-Image place recognition tasks.
\subsubsection{Text-LiDAR place recognition}

\begin{table*}[t]
    \centering
    \caption{Performance comparison for Image-LiDAR place recognition on the KITTI-360 dataset.}
    \label{imagepc}
    \resizebox{0.80\textwidth}{!}{%
    \begin{tabular}{@{}lccc|ccc|ccc|ccc@{}}
        \toprule
         & \multicolumn{6}{c}{LiDAR-to-Image} & \multicolumn{6}{c}{Image-to-LiDAR} \\ 
        \cmidrule(lr){2-7} \cmidrule(lr){8-13}
        & \multicolumn{3}{c}{val} & \multicolumn{3}{c}{test} & \multicolumn{3}{c}{val} & \multicolumn{3}{c}{test} \\ 
        Methods & R@1 & R@3 & R@5 & R@1 & R@3 & R@5
        & R@1 & R@3& R@5 & R@1 & R@3 & R@5 \\ \midrule
        LIP-Loc  \cite{shubodh2024lip}  & 48.2 & 61.8 & 66.7 & 54.0 & 67.8 & 75.1   & 49.4 & 62.5 & 70.9 & 49.3 & 64.1 & 70.9  \\ 
        Cattaneo\cite{cattaneo2020global}    & 11.2 & 17.6 & 21.3 & 21.8 & 28.4 & 33.5   & 18.8 & 31.9 & 38.9 & 25.3 & 36.8 & 43.7 \\ 
        VXP \cite{li2024vxp}   & 33.5 & 37.3 & 40.8 & 19.7 & 26.1 & 29.1   & 43.5 & 55.5 & 61.8 & 31.9 & 42.7 & 48.0 \\ 
        UniLoc (Ours)  & \textbf{91.8} & \textbf{95.7} & \textbf{96.3} & \textbf{94.4} & \textbf{97.5} & \textbf{98.0} & \textbf{92.0} & \textbf{96.0} & \textbf{97.4} & \textbf{93.5} & \textbf{96.5} & \textbf{97.7} \\ 
        \bottomrule
    \end{tabular}%
        }
\end{table*}

\begin{table*}[htbp!]
    \centering
    \caption{Performance comparison for Image-Image and LiDAR-LiDAR place recognition on the KITTI-360 dataset.}
    \label{same modality}
    \resizebox{0.80\textwidth}{!}{%
    \begin{tabular}{@{}lccc|ccc|ccc|ccc@{}}
        \toprule
         & \multicolumn{6}{c}{Image-to-Image} & \multicolumn{6}{c}{LiDAR-to-LiDAR} \\ 
        \cmidrule(lr){2-7} \cmidrule(lr){8-13}
        & \multicolumn{3}{c}{val} & \multicolumn{3}{c}{test} & \multicolumn{3}{c}{val} & \multicolumn{3}{c}{test} \\ 
        Methods & R@1 & R@3 & R@5 & R@1 & R@3 & R@5
        & R@1 & R@3 & R@5 & R@1 & R@3 & R@5 \\ \midrule
        LIP-Loc~\cite{shubodh2024lip}       & 85.3 & 93.2 & 95.3 & 87.7 & 94.1 & 96.0 & 85.7 & 92.3 & 93.9 & 82.2 & 89.8 & 92.4 \\ 
        Cattaneo's~\cite{cattaneo2020global}      & 100 & 100 & 100 & 99.9 & 99.9 & 100 & 95.6 & 96.3 & 96.5 & 90.5 & 93.8 & 94.6 \\ 
        VXP~\cite{li2024vxp}      & 100 & 100 & 100 & 99.9 & 99.9 & 99.9 & 98.3 & 98.8 & 99.3 & 96.5 & 97.8 & 98.2\\ 
        CASSPR~\cite{Xia_2023_ICCV}       & - & - & - & - & - & - & 98.8 & 99.3 & 99.7 & 98.1 &99.3& 99.4 \\
        AnyLoc\cite{keetha2023anyloc}   & 100& 100 & 100 & 99.8 & 99.9 & 99.9 & - & - & - & - & - & - \\

        MixVPR\cite{ali2023mixvpr} & 100 & 100 & 100 & 99.9 & 99.9 & 100 & - & - & - & - & - & - \\
        UniLoc (Ours)  & 96.7 & 99.3 & 99.7 & 96.5 & 98.5 & 98.9 & 96.9 & 98.8 & 99.0 & 97.2 & 99.1 & 99.2 \\ 
        \bottomrule
    \end{tabular}%
        }
\end{table*}

\begin{figure}[!htbp]
    \centering
    \includegraphics[width=\linewidth]{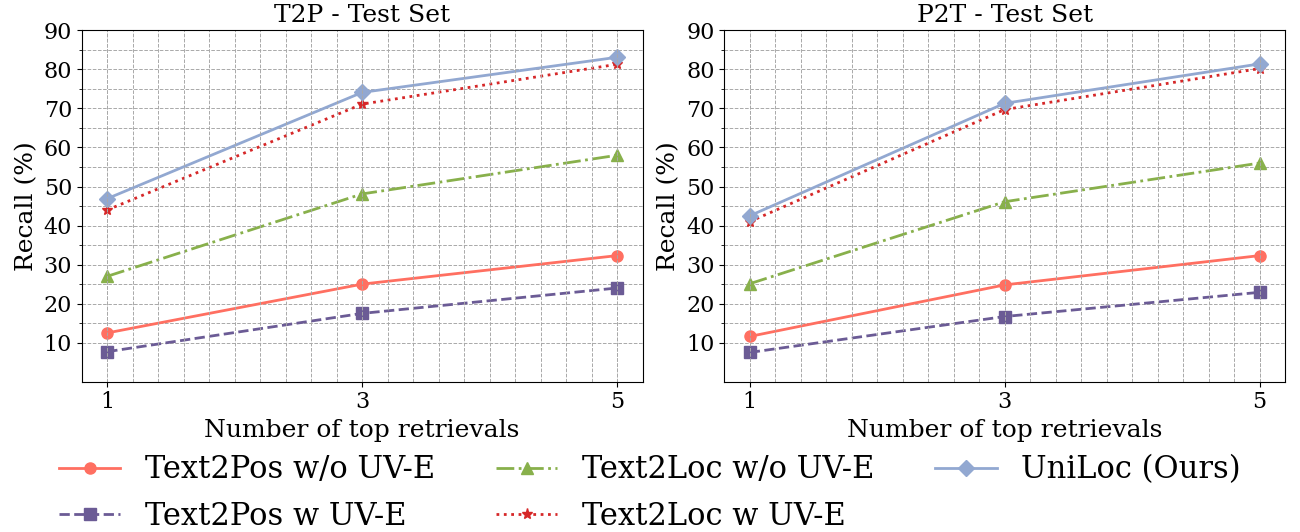}
    \caption{Performance comparison for Text-LiDAR place recognition on the KITTI-360 dataset. “w/o UV-E” indicates removing the UV encoder in the instance encoder and the instance position descriptor in the text, while “w UV-E” indicates the opposite setup.}
    \label{fig:image-pc}
\end{figure}

We compare our UniLoc with state-of-the-art Text-LiDAR methods Text2Pos~\cite{kolmet2022text2pos} and Text2Loc~\cite{xia2024text2loc}, both with and without our UV Encoder. As shown in Fig. \ref{fig:image-pc}, UniLoc achieves the best performance across all recall at top-1/3/5. For Text-to-Lidar retrieval, UniLoc demonstrates notable improvements, surpassing the best-performing Text2Loc by 2.9\%  at top-1 on the test set. Moreover, for Lidar-to-Text retrieval, UniLoc outperforms other methods, with improvements of 1.5\% at top-1 on the test set compared to the Text2Loc w/UV Encoder configuration. These results underscore UniLoc's strong retrieval performance, demonstrating its robustness and effectiveness in handling Text-to-Lidar retrieval tasks.

\subsubsection{Image-LiDAR place recognition}
We compare our UniLoc with state-of-the-art Image-LiDAR models, including LIP-Loc~\cite{shubodh2024lip}, Cattaneo's~\cite{cattaneo2020global}, and VXP~\cite{li2024vxp}. As shown in Table \ref{imagepc}, UniLoc significantly improves over existing models in the Image-LiDAR place recognition task across all recall levels. For LiDAR-to-Image retrieval, UniLoc outperforms other methods by achieving gains of 43.6\%, 33.9\%, and 29.6\% at top-1/3/5 on the validation set, and 40.4\%, 29.7\%, and 22.9\% on the test set, when compared to the best-performing baseline, LIP-Loc. 
For Image-to-LiDAR retrieval, UniLoc demonstrates notable advancements, surpassing LIP-Loc with increases of 42.6\% and 44.2\% at top-1 on the validation and test set, respectively. These marked improvements underscore UniLoc’s superior retrieval capabilities, highlighting its robustness and accuracy in Image-LiDAR retrieval tasks.

\subsubsection{Image-based and LiDAR-based place recognition}
We further compare UniLoc with several state-of-the-art uni-modal models, including 3D-3D model CASSPR~\cite{Xia_2023_ICCV} and 2D-2D models AnyLoc~\cite{keetha2023anyloc} and MixVPR~\cite{ali2023mixvpr}. Additionally, we compare with leading 2D-3D models  VXP~\cite{li2024vxp}, Cattaneo's~\cite{cattaneo2020global}, and LIP-Loc~\cite{shubodh2024lip}. As shown in Table \ref{same modality}, UniLoc achieves competitive performance in Image-to-Image and LiDAR-to-LiDAR retrieval tasks.
For image-based place recognition, UniLoc achieves the top-1 recall rate of 96.5 on the test set. Additionally, UniLoc reaches 97.2 recall at top-1 for LiDAR-based place recognition. That demonstrates that UniLoc achieves the better performance compared to the previous cross-modal methods.

\subsection{Ablation study}
In this section, we analyze the hyperparameter selection in scene-level matching and evaluate the effectiveness of various components of UniLoc. The results in Table \ref{component}  are obtained by averaging the performance on the Text-Image and Image-LiDAR place recognition. All experiments are conducted using 2D ground truth segmentation labels.


\noindent \textbf{Effectiveness of various components.}
To validate the effectiveness of each component in UniLoc, we conduct a series of ablation studies: replacing self-attention based pooling with max pooling in the SAP module (as presented in Section \ref{scene model}), removing the UV encoder (presented in Section \ref{instance block}) and the instance position descriptor in text, not loading the pre-trained instance-level model during UniLoc training, and substituting the CLIP image and text encoders with DINO \cite{caron2021emerging} and T5 \cite{raffel2020exploring}. All networks are trained on the KITTI-360 dataset, with results summarized in Table \ref{component}. The experimental results indicate that the self-attention based pooling module enhances the model’s average recall by 2.7\% on the validation set and 1\% on the test set. Furthermore, incorporating object location descriptions in the text and using the UV encoder increases the model’s average recall by 15.8\% on the validation set and 13.9\% on the test set. Additionally, the findings reveal that loading the pre-trained instance-level network improves the model’s average recall by 2.3\% on the validation set and 2.9\% on the test set. Due to the superior alignment capabilities between images and text in CLIP, compared to DINO+T5, using CLIP enhances the model’s average recall by 6.7\% on the validation set and 5.7\% on the test set.

\begin{table}[t]
    \centering
\caption{Ablation study of our UniLoc on the KITTI-360 dataset. ”w/o SAP” indicates replacing the proposed self-attention based pooling (SAP) module with the max-pooling layer. ”w/o UV-E” indicates removing all encoding modules related to UV coordinates. ”w/o pre-trained” indicates not loading the pre-trained weights from instance-level matching stage.
”w/o CLIP” indicates replacing the pre-trained CLIP image and text encoder with DINO~\cite{zhang2022dino} and T5~\cite{raffel2020exploring}.}
    \label{component}
    \resizebox{0.45\textwidth}{!}{%
    \begin{tabular}{@{}lcccc|cccc@{}}
        \toprule
         & \multicolumn{4}{c}{val} & \multicolumn{4}{c}{test} \\ 
        \cmidrule(lr){2-5} \cmidrule(lr){6-9}
        Methods & R@1 & R@3 & R@5 & Ave & R@1 & R@3 & R@5 & Ave \\ \midrule
        w/o SAP       & 61.7 & 86.3 & 92.9   & 80.3 & 73.7 & 94.3 & 97.6 & 88.5 \\ 
        w/o UV-E       & 46.9 & 72.9 & 81.9 & 67.2 & 57.6 & 81.5 & 87.8 & 75.6 \\ 
        w/o pre-trained       &62.7 & 86.6 & 92.8 & 80.7 & 71.3& 92.4 & 96.2 & 86.6 \\ 
        w/o CLIP     & 56.7 & 82.2 & 90.1 & 76.3 & 67.0 & 89.9 & 94.6 & 83.8 \\ 
        UniLoc (Ours) & \textbf{65.6} & \textbf{89.1} & \textbf{94.3} & \textbf{83.0} & \textbf{75.5} & \textbf{95.1} & \textbf{97.9} & \textbf{89.5} \\ 
        \bottomrule
    \end{tabular}%
    }
\end{table}


\subsection{Computational cost analysis}
In this section, we analyze the required computational resources of our UniLoc regarding the number of parameters and time efficiency. 
As shown in Table \ref{computational cost}, the total parameter count of UniLoc (excluding the parameters from the pre-trained CLIP image and text encoders) is \SI{22.5}{M}. 
UniLoc comprises \SI{12.1}{M}, \SI{3.4}{M}, and \SI{7.1}{M} parameters for the point cloud, text, and image branches, respectively.
Using a single NVIDIA A40 (48G) GPU, the point cloud instance block in UniLoc requires \SI{2.2}{ms} to compute a 3D instance-level descriptor and \SI{0.8}{ms} to aggregate them into a scene-level descriptor. For text, it takes \SI{0.082}{ms} for instance-level computation and \SI{0.59}{ms} for aggregation. The image instance block takes \SI{12.6}{ms} for instance-level computation and \SI{0.66}{ms} for aggregation.


\begin{table}[h!]
\centering
\caption{Computational cost requirement analysis of our UniLoc on the KITTI-360 test dataset.}
\label{computational cost}
\resizebox{0.45\textwidth}{!}{%
\begin{tabular}{@{}lccccc@{}}
\toprule
\multicolumn{1}{l}{} & \multicolumn{3}{c}{Parameters (M)} & \multicolumn{2}{c}{Time Usage (ms)} \\ \cmidrule(lr){2-4} \cmidrule(l){5-6}
Branch       & Total & Instance Block & SAP & Instance Descriptor & Scene Descriptor \\ \midrule
Point Cloud  & 12.1  & 8.7              & 3.4  & 2.2                  & 0.8              \\
Image        & 7.1   & 3.7              & 3.4  & 12.6                 & 0.66             \\
Text         & 3.4   & 0.8              & 2.6  & 0.082                & 0.59             \\ \bottomrule
\end{tabular}%
}
\label{tab:parameters_and_times}
\end{table}

\subsection{Embedding space analysis}
\begin{figure}[htbp]
    \centering
    \includegraphics[width=1\linewidth]{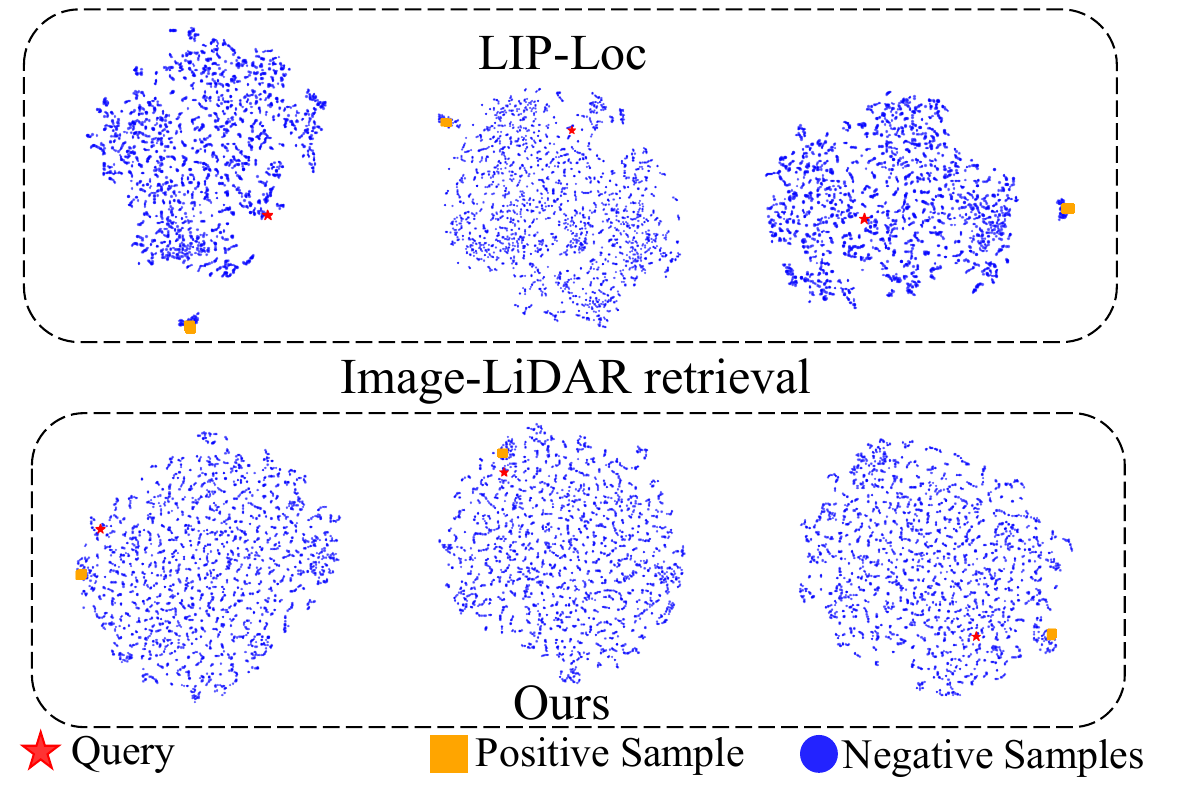}
    \caption{T-SNE visualization for the Image-LiDAR recognition.}
    \label{fig:tsne}
\end{figure}

We employ T-SNE~\cite{van2008visualizing} to visually represent the learned embedding space, as illustrated in Fig. ~\ref{fig:tsne}. The baseline method LIP-Loc~\cite{shubodh2024lip} produces a less discriminative embedding space, where positive LIDAR samples are often far from the query images and scattered throughout the space. In contrast, our UniLoc brings positive point cloud samples and query images significantly closer within the embedding space, demonstrating that UniLoc creates a more discriminative cross-modal space for place recognition.


%% file: sec/7_Conclusion.tex
\section{Conclusion}
We have presented UniLoc, a method for cross-modal place recognition that can also operate competitively on a single modality at a time. Our experiments show that factoring the place recognition problem into instance and scene levels for different modalities is an effective solution for place recognition.
We hope that cross-modal place recognition results in more robust systems in the future, capable of coping with the loss of different sensors (e.g. under adversarial lighting conditions) and locating a place via natural language descriptions seamlessly.
Future work could include explorations of other modalities, such as sound, infrared or even event cameras.


%% file: sec/X_suppl.tex
\clearpage
\setcounter{page}{1}
\maketitlesupplementary


\noindent In this Supplementary Material, we present extensive additional experiments conducted on the KITTI-360 dataset, further illustrating the effectiveness of UniLoc. Additionally, we provide detailed insights into the experimental setup and dataset. In Sec. \ref{dataset details}, we discuss the dataset generation process and its key specifics. Sec. \ref{Implement details} highlights implementation details, clarifying the methods employed. In Sec. \ref{qualitative ana}, we offer a qualitative analysis of the top-3 candidate retrieved results, demonstrating their place recognition performance across diverse tasks. Lastly, Sec. \ref{more ablation} presents additional ablation studies to rigorously validate the robustness and general effectiveness of our UniLoc. \textbf{We also provide our code for reproducibility in this Supp. Submission.}

\section{Dataset Details} \label{dataset details}
\begin{figure}
    \centering
    \includegraphics[width=1\linewidth]{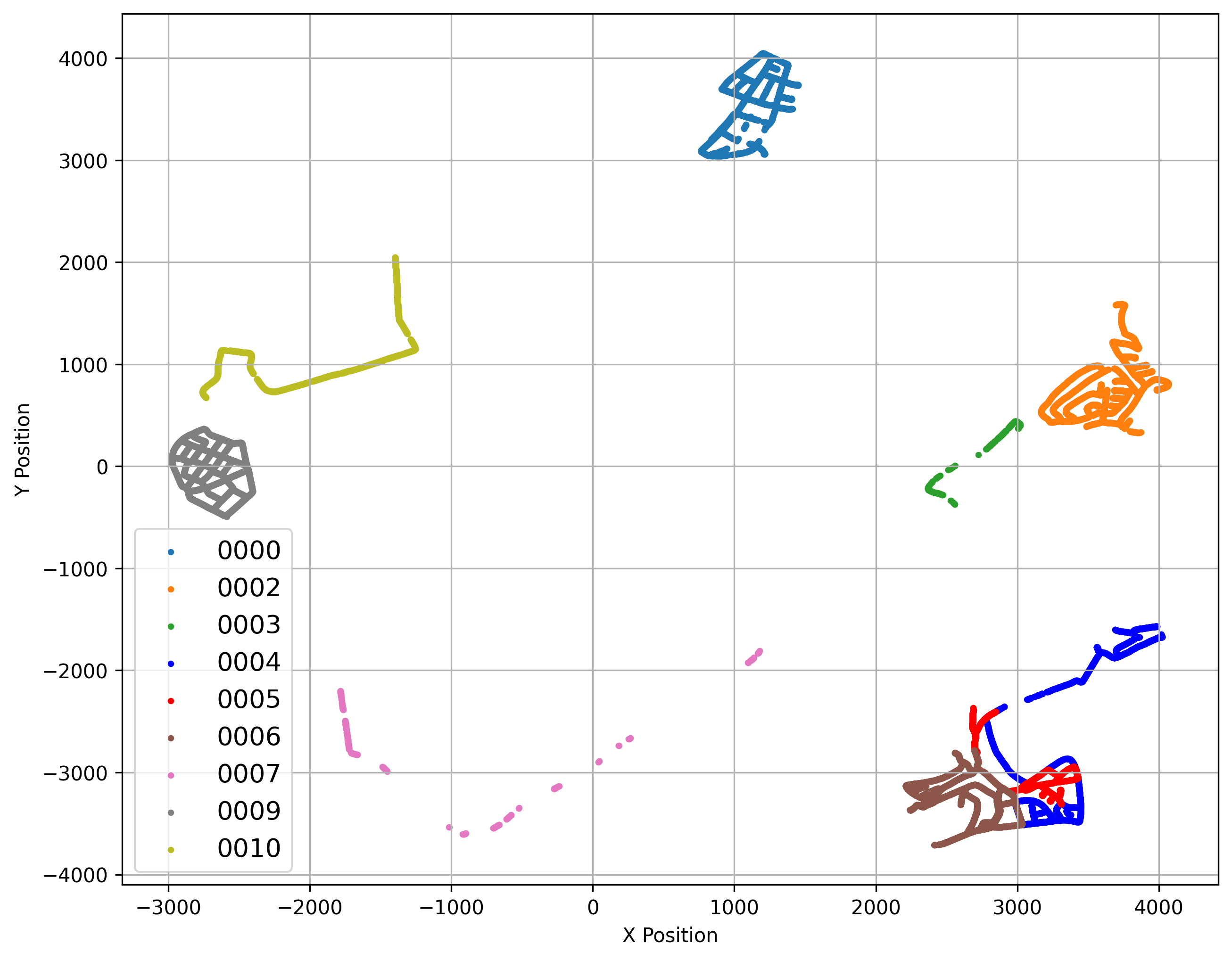}
    \caption{The trajectory distribution of the processed KITTI-360 dataset.}
    \label{fig:dataset trajectory}
\end{figure}

\begin{figure}
    \centering
    \includegraphics[width=1\linewidth]{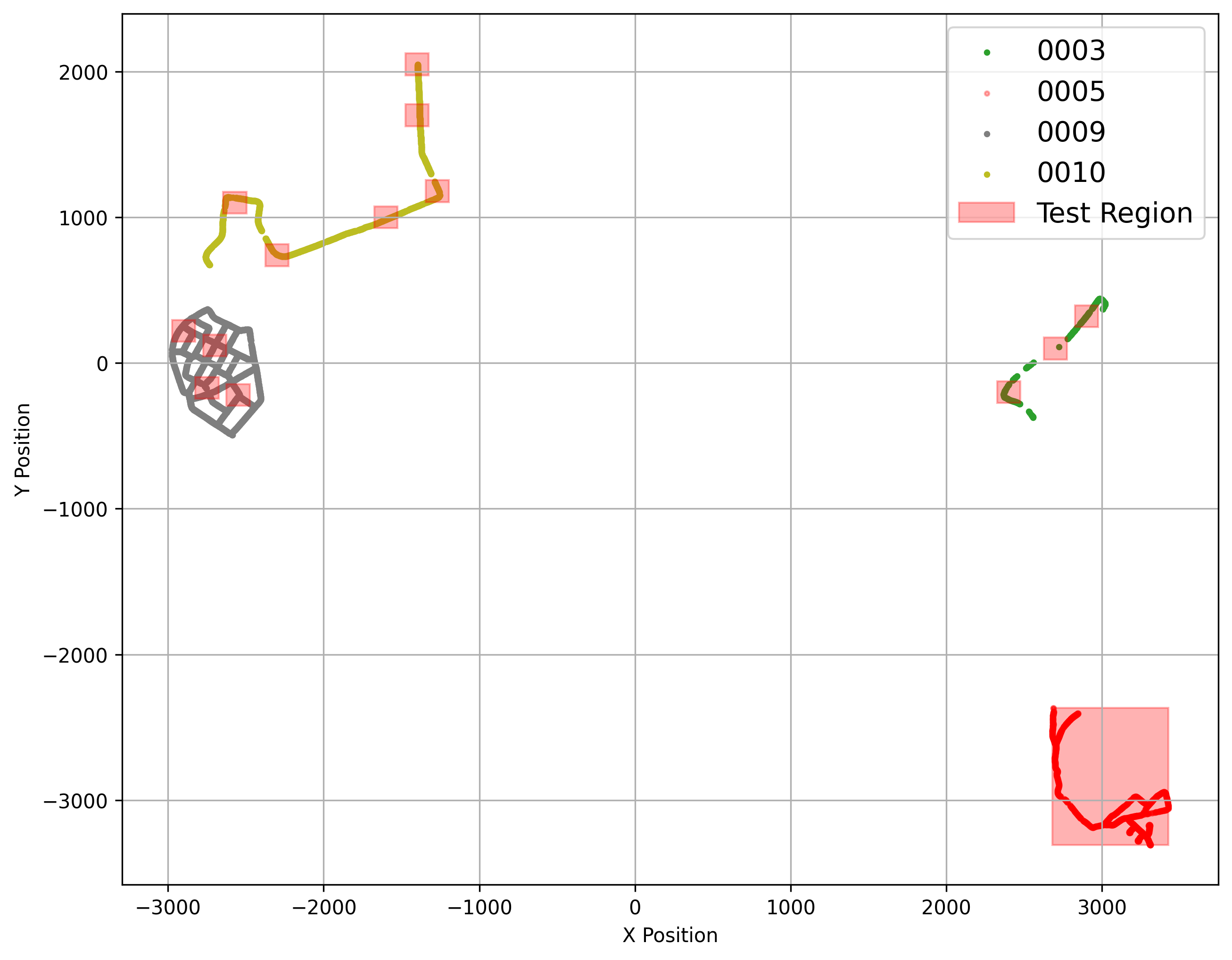}
    \caption{The trajectory distribution of the validation and test sets in the processed KITTI-360 dataset. The red rectangles represent the defined test regions.}
    \label{fig:val trajectory}
\end{figure}

\paragraph{Dataset trajectory distribution.} To ensure a one-to-one correspondence between the textual descriptions and the image and point cloud data in the place recognition tasks, we deduplicated the dataset using text descriptions generated from the images, maintaining a minimum distance of one meter between scenes. Fig. \ref{fig:dataset trajectory} depicts the spatial distribution of trajectories across the dataset, while Fig. \ref{fig:val trajectory} highlights the distribution of test and validation data used in the experiments for Tab. 1 and 2. (in main paper). A test region refers to a specific area within the dataset designated for generating queries to evaluate the model's retrieval performance. For sequences 03, 09, and 10, distinct test regions were defined, and queries were restricted to these areas. In the case of sequence 05, its close proximity to the training data from sequences 04 and 06 made it challenging to define non-overlapping test regions. As a result, every data point in sequence 05 is treated as a query, effectively treating the entire sequence as a test region.

\paragraph{Pre-processing of point cloud.} Most automotive datasets focus primarily on the instance segmentation of dynamic objects such as vehicles and pedestrians. In contrast, KITTI360 expands its annotation scope to include static object categories like buildings and traffic lights. These static objects provide consistent and reliable references, making them particularly valuable for localization tasks. Similar to \cite{kolmet2022text2pos}\cite{xia2024text2loc} ,in this work, we exploit static object instances not only to generate position queries but also as essential cues to improve the accuracy and robustness of position localization. We utilize the semantic IDs in the KITTI-360 dataset's Accumulated Point Clouds to filter the desired object classes. Using instance IDs, we further distinguish and aggregate all instances within each class into individual objects. At this stage, each object in the point cloud is assigned a unique and consistent instance ID. However, this processing results in numerous objects for each class. The purpose of this step is to aggregate objects and downsample them according to their class. For instance, buildings are downsampled using a voxel size of 0.25, while poles are not downsampled. This approach ensures that objects from different classes maintain a roughly consistent number of points. Additionally, we filter out objects with fewer points than a class-specific threshold (e.g., buildings require at least 250 points, while poles require a minimum of 25 points). Subsequently, we sample the processed Accumulated Point Clouds using a 40-meter radius around the positions of images in the dataset. During this process, smaller objects that lose more than one-third of their points due to sampling are discarded. This decision is made because PointNet++ \cite{qi2017pointnet} struggles to capture the semantic information of small objects (e.g., lamps, trash bins) with significant point cloud loss. Notably, to ensure the average color of the point clouds aligns with the textual descriptions derived from the corresponding images, the RGB color of each point is assigned based on the pixel value at its projected UV coordinates in the image.  The remaining steps of the processing pipeline have already been described in detail in the main paper.

\noindent Each scene is designed to include a fixed 12 instances. For scenes containing more than 12 instances, instances are randomly filtered to retain only 12. Notably, all text-related experiments (Text-to-LiDAR, LiDAR-to-Text, Text--to-Image, and Image-to-Text) utilize text generated directly from the images. Furthermore, because it is visually challenging to distinguish between "pole" and "small pole," these two classes are merged into a single "pole" category across all three modalities in the dataset.

\section{Implementation Details} \label{Implement details}

\paragraph{Training parameters.}
We train our UniLoc using the Adam optimizer for both instance-level and scene-level matching stages, initializing the learning rate (LR) at 5e-4. The instance pre-training phase consists of training the model for 20 epochs with a batch size of 256. Subsequently, the final model undergoes an additional 20 epochs of training with a batch size of 64. A multi-step training schedule is employed for both instance-level and scene-level training, where the LR is reduced by a factor of 0.4 every 7 epochs. The temperature coefficient  \( \tau \) is fixed at 0.1. To encode individual instances within the scene, we leverage PointNet++ \cite{qi2017pointnet} as implemented in \cite{kolmet2022text2pos}.\

\paragraph{SAP details.} For the Multi-Head Self-Attention (MHSA) used in SAP (as described in Section. 4.2 in main paper), the text branch utilizes a 4-head, 1-layer MHSA, while the point cloud and image branches employ a 4-head, 2-layer MHSA. Furthermore, the Pooling block of Self-Attention-based Pooling (SAP) module for each branch is implemented using a three-layer MLP with a parameter size of \SI{0.13}{M}.\

\paragraph{Experimental details.} To train and test the model with textual descriptions, we incorporate a tolerance mechanism for scene descriptions. Instead of relying on the descriptions of all instances within a scene, six sentences are randomly selected as the scene's text description during both training and testing. This design reflects real-world human behavior, where it is rare to comprehensively describe every instance in a scene without omissions. During training, six sentences are randomly sampled from the descriptions of all instances in a scene to ensure the full utilization of the textual dataset. For testing, the same random seed is used to maintain consistency in results.\\
For the Image-to-Text task, to ensure a fair comparison, ground-truth semantic masks are excluded during the evaluation phase. Instead, we train a PSPNet \cite{zhao2017pyramid} model, equipped with a ResNet101 \cite{he2016deep} backbone, using paired semantic masks and raw images. PSPNet model (or an equivalent advanced segmentation model) is employed to segment the validation and test sets. During the evaluation of UniLoc, the semantic masks produced by PSPNet are used to segment the raw images, serving as input to the image branch.

\section{Qualitative Analysis}
\begin{figure*}[!htbp]
    \centering
    \includegraphics[width=\linewidth]{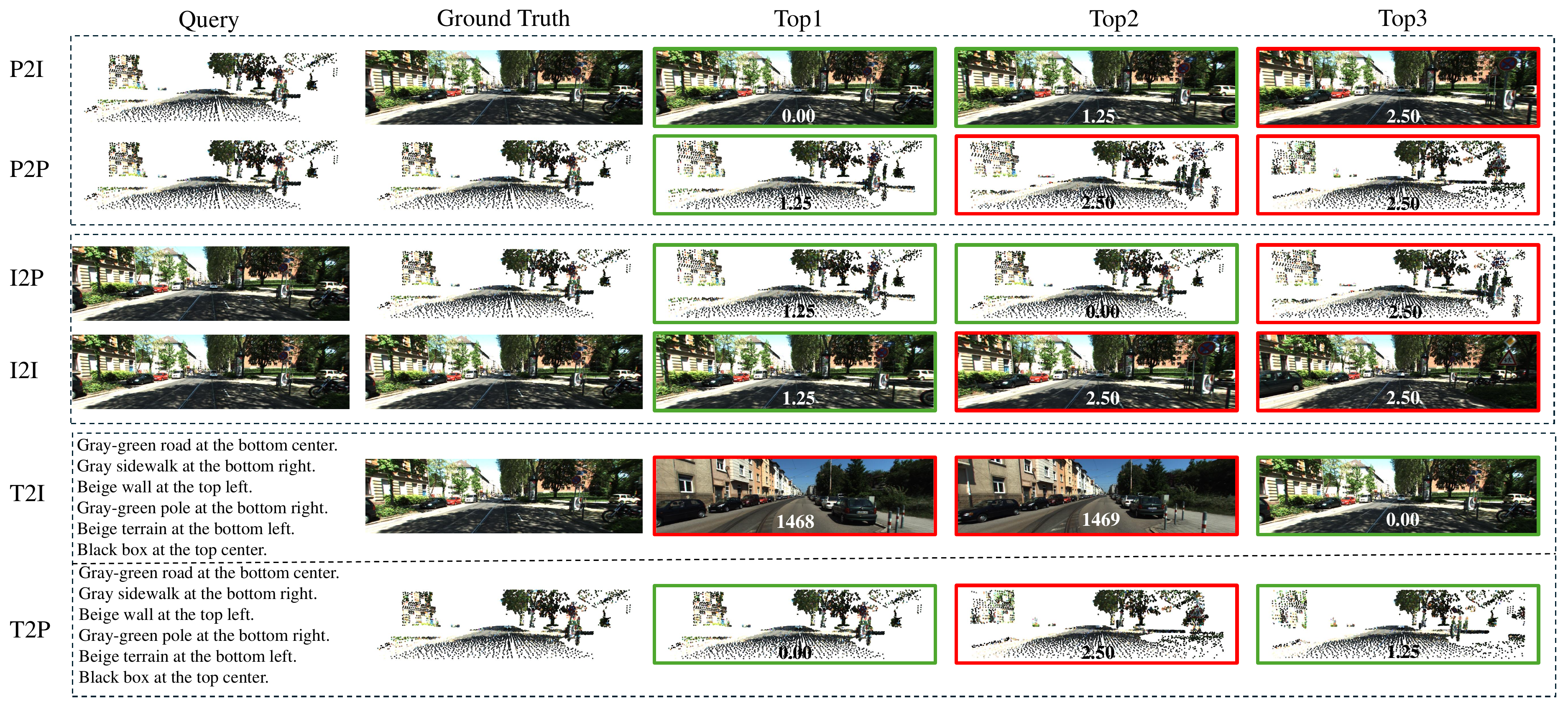}
    \caption{Qualitative localization results on the KITTI-360 dataset: In the context of place recognition, the numbers on the top-3 retrieved results indicate the center distances between the retrieved results and the ground truth. The results that were retrieved within 2 meters of the query are highlighted in green, signifying successful localization. Retrievals exceeding this distance are marked in red. Here, P2I denotes LiDAR-to-Image, P2P: LiDAR-to-LiDAR, I2P: Image-to-LiDAR, I2I: Image-to-Image, T2I: Text-to-Image, and T2P: Text-to-LiDAR tasks.}
\label{fig:Qualitative analysis}
\end{figure*}

In addition to the quantitative results, we present qualitative examples in Fig. \ref{fig:Qualitative analysis}, showcasing the place recognition performance among images, point clouds, and corresponding text sampled from the same location. Given any one of these three modalities, UniLoc allows for the retrieval of data from either the same or a different modality. In Fig. \ref{fig:Qualitative analysis}, we visualize the results for six tasks: LiDAR-to-Image(P2I), LiDAR-to-LiDAR(P2P), Image-to-LiDAR(I2P), Image-to-Image(I2I), Text-to-Image(T2I), and Text-to-LiDAR(T2P). 
For each task, we display the ground truth of the target modality (the ground truth of target modality sampled from the same location as the query) alongside the top-3 retrieved results. 
It is worth noting that for I2I and P2P tasks, the query itself is removed from the database, ensuring that the query cannot retrieve itself.

In most cases, UniLoc successfully retrieves target modality data sampled near the query, demonstrating its strong performance in cross-modal place recognition tasks. UniLoc also fails to locate target modality data close to the query in the Top-1 and Top-2 results, as shown in Fig. \ref{fig:Qualitative analysis} (T2I). 
We find that even when the positions of the retrieval results are far from the ground truth, they often contain instances that are visually or semantically similar to them. This highlights the challenge posed by the low diversity of outdoor scenes and suggests that richer and more accurate textual descriptions could help mitigate ambiguity in target modality representations, thereby enhancing localization accuracy.

\label{qualitative ana}
\section{More Ablation Studies}
\label{more ablation}
In this section, we show more experimental results and ablation studies.

\paragraph{Selection of \( \alpha \).} In Tab. ~\ref{lamda}, we examine the impact of different \( \alpha \) values in Eq. 9 (in main paper) on model performance. The results indicate that the model achieves optimal performance on both the test and validation datasets when \( \alpha \) is set to 0.3.
\begin{table}[!htbp]
    \centering
\caption{Hyperparameter analysis in the scene-level matching loss.}
    \label{lamda}
\resizebox{0.45\textwidth}{!}{%
    \begin{tabular}{@{}lcccc|cccc@{}}
        \toprule
         & \multicolumn{4}{c}{val} & \multicolumn{4}{c}{test} \\ 
        \cmidrule(lr){2-5} \cmidrule(lr){6-9}
        $\alpha$ & R@1 & R@3 & R@5 & Ave & R@1 & R@3 & R@5 & Ave \\ \midrule
        0.1       & 62.7 & 86.1 & 92.6 & 80.5 & 71.0 & 91.4 & 95.3 & 85.9 \\ 
        0.3       & \textbf{65.6} & \textbf{89.1} & \textbf{94.3} & \textbf{83.0} & \textbf{75.5} & \textbf{95.1} & 97.9 & \textbf{89.5} \\ 
        0.5       & 65.4 & 88.8 & 93.9 & 82.7 & \textbf{75.5} & \textbf{95.1} & \textbf{98.0} & \textbf{89.5} \\ 
        0.7       & 63.3 & 88.0 & 93.9 & 81.7 & 73.5 & 94.4 & 97.6 & 88.5 \\ 
        0.9       & 63.1 & 86.8 & 92.8 & 80.9 & 74.3 & 94.5 & 97.7 & 88.8 \\ 
        \bottomrule
    \end{tabular}%
        }
\end{table}\\

\paragraph{Robustness analysis for the number of textual descriptions.}
As shown in Tab. \ref{text image text} and \ref{text pc text}, we explored the impact of textual description richness on model performance. In this experiment, we evaluated the performance of the Text-to-Image and Text-to-LiDAR place recognition using language descriptions containing 4, 5, and 6 sentences, where each sentence corresponds to the description of a single instance. Notably, the size of the textual descriptions used during training remained fixed at six sentences. The Text-to-Image place recognition was tested using ground-truth segmentation masks.
We observed that even with text descriptions limited to only 6 sentences, our method achieves a recall performance of 72. 5\% at top-1 in the KITTI-360 test set, demonstrating its effectiveness. We also believe that adding more descriptive information will further improve performance.
Moreover, the findings show that reducing the textual input to five sentences does not significantly impact performance, demonstrating that our UniLoc is robust to a certain extent of missing textual descriptions.

\begin{table}[!htbp]
    \centering
\caption{Performance comparison of Text-to-Image place recognition on the KITTI-360 dataset with different numbers of instance-level textual descriptions.}
    \label{text image text}
\resizebox{0.45\textwidth}{!}{%
    \begin{tabular}{@{}lcccc|cccc@{}}
        \toprule
         & \multicolumn{4}{c}{val} & \multicolumn{4}{c}{test} \\ 
        \cmidrule(lr){2-5} \cmidrule(lr){6-9}
        \ Num. & R@1 & R@3 & R@5 & Ave & R@1 & R@3 & R@5 & Ave \\ \midrule
        6       & 65.8 & 89.1 & 94.5 & 83.1 & 72.5 & 93.6 & 97.2 & 87.8 \\ 
        5       & 49.6 & 75.5 & 85.7 & 70.3 & 56.7 & 82.1 & 89.3 & 76.0 \\ 
        4       & 29.9 & 54.9 & 67.3 & 50.7 & 29.3 & 52.0 & 62.5 & 47.9 \\ 
        \bottomrule
    \end{tabular}%
        }
\end{table}

\begin{table}[!htbp]
    \centering
\caption{Performance comparison of Text-to-LiDAR place recognition on the KITTI-360 dataset with different numbers of instance-level textual descriptions.}
    \label{text pc text}
\resizebox{0.45\textwidth}{!}{%
    \begin{tabular}{@{}lcccc|cccc@{}}
        \toprule
         & \multicolumn{4}{c}{val} & \multicolumn{4}{c}{test} \\ 
        \cmidrule(lr){2-5} \cmidrule(lr){6-9}
        \ Num. & R@1 & R@3 & R@5 & Ave & R@1 & R@3 & R@5 & Ave \\ \midrule
        6       & 36.4 & 62.2 & 73.5 & 57.4 & 46.7 & 73.6 & 83.3 &67.9 \\ 
        5       & 24.8 & 46.8 & 58.6 & 43.4 & 30.9 & 55.1 & 65.7 & 50.6 \\ 
        4       & 12.2 & 27.9 & 36.9 & 25.7 & 12.5 & 26.5 & 35.5 & 24.8  \\ 
        \bottomrule
    \end{tabular}%
        }
\end{table}

\begin{table*}[!htbp]
    \centering
    \caption{Performance comparison of Image-LiDAR place recognition on the KITTI-360 dataset under different distance thresholds}
    \label{ablation image-pc}
    \resizebox{0.9\textwidth}{!}{%
    \begin{tabular}{@{}lccc|ccc|ccc|ccc@{}}
        \toprule
         & \multicolumn{6}{c}{LiDAR-to-Image} & \multicolumn{6}{c}{Image-to-LiDAR} \\ 
        \cmidrule(lr){2-7} \cmidrule(lr){8-13}
        & \multicolumn{3}{c}{val} & \multicolumn{3}{c}{test} & \multicolumn{3}{c}{val} & \multicolumn{3}{c}{test} \\ 
        $d$ (m) & R@1 & R@3 & R@5 & R@1 & R@3 & R@5
        & R@1 & R@3 & R@5 & R@1 & R@3 & R@5 \\ \midrule
        20    & 95.4 & 98.9 & 99.4 & 99.0 & 99.8 & 99.9   & 95.0 & 98.7 & 99.4 & 98.9 & 99.7 & 99.9  \\ 
        10    & 93.3 & 98.5 & 99.2 & 98.8 & 99.7 & 99.9   & 92.6 & 97.9 & 98.8 & 98.5 & 99.6 & 99.8 \\ 
        5     & 88.7 & 96.7 & 98.6 & 97.9 & 99.6 & 99.8   & 87.9 & 96.6 & 97.9 & 97.7 & 99.6 & 99.8 \\ 
        2     & 80.0 & 94.0 & 97.2 & 93.1 & 99.1 & 99.6   & 78.4 & 93.1 & 96.2 & 93.6 & 99.3 & 99.7 \\ 
        \bottomrule
    \end{tabular}%
    }
\end{table*}

\begin{table*}[!htbp]
    \centering
    \caption{Performance comparison of Text-Image place recognition on the KITTI-360 dataset under different distance thresholds}
    \label{ablation image-text}
    \resizebox{0.9\textwidth}{!}{%
    \begin{tabular}{@{}lccc|ccc|ccc|ccc@{}}
        \toprule
         & \multicolumn{6}{c}{Text-to-Image} & \multicolumn{6}{c}{Image-to-Text} \\ 
        \cmidrule(lr){2-7} \cmidrule(lr){8-13}
        & \multicolumn{3}{c}{val} & \multicolumn{3}{c}{test} & \multicolumn{3}{c}{val} & \multicolumn{3}{c}{test} \\ 
        $d$ (m) & R@1 & R@3 & R@5 & R@1 & R@3 & R@5
        & R@1 & R@3 & R@5 & R@1 & R@3 & R@5 \\ \midrule
        20    & 93.7 & 98.4 & 99.6 & 94.5 & 98.5 & 99.3   & 93.0 & 98.4 & 99.2 & 94.5 & 98.5 & 99.1 \\ 
        10    & 90.7 & 97.9 & 99.2 & 93.9 & 98.2 & 99.1   & 90.1 & 97.8 & 98.9 & 93.7 & 98.1 & 99.0 \\ 
        5     & 86.5 & 96.5 & 98.7 & 91.9 & 97.8 & 98.9   & 85.2 & 97.1 & 98.3 & 91.4 & 97.6 & 98.7 \\ 
        2     & 78.3 & 94.0 & 97.2 & 85.9 & 96.7 & 98.4   & 76.1 & 93.5 & 96.5 & 84.6 & 96.2 & 98.2 \\ 
        \bottomrule
    \end{tabular}%
    }
\end{table*}

\paragraph{Place recognition under different distance thresholds.}
In this ablation study, we regard the retrieved result as a correct match if the distance is within $d$.
To evaluate UniLoc's cross-modal place recognition capabilities between images and LiDAR (Image-to-LiDAR, LiDAR-to-Image tasks), we set the $d$ at four different values ranging from \SI{20}{m} to \SI{2}{m}. 
Note that we do not exclude cross-modal data from the same sampling location as the query in the database. As shown in Tab. \ref{ablation image-pc}, Image-to-LiDAR place recognition achieves the best performance not only at the $d=20m$ but also maintains a good recall rate even under the strict \SI{2}{m} threshold.

\noindent For Image-to-Text and Text-to-Image place recognition, we conducted experiments under the same conditions, as shown in Tab \ref{ablation image-text}. These results indicate that our UniLoc has strong cross-modal generalization capabilities across different input types under different distance thresholds.

\paragraph{Comparison with different backbones used in scene-level matching.}
To evaluate whether the pre-trained CLIP\cite{radford2021learning} is the most suitable backbone, we also experimented with replacing the CLIP with the BLIP\cite{li2022blip} to encode text and images. Notably, we do not utilize any pre-trained instance-level models here. Follow the same methodology as described in \textbf{Section 6.3} of the main paper, where performance is averaged across Text-Image (Text-to-Image and Image-to-Text tasks) and Image-LiDAR (Image-to-LiDAR and LiDAR-to-Image tasks) place recognition. All experiments were conducted using 2D ground-truth semantic labels. The results are presented in Tab. \ref{backbone}. We can see that upgrading the backbone from CLIP to a more advanced model like BLIP can significantly enhance UniLoc's overall performance. Therefore, we believe that integrating even more powerful models, such as BLIP2~\cite{li2023blip}, could further boost its capabilities. This finding also highlights our UniLoc can fully leverage recent advances in large-scale contrastive learning.


\begin{table}[!htbp]    
    \centering
\caption{Performance comparison with different backbones used in scene-level matching.}
    \label{backbone}
    \resizebox{0.45\textwidth}{!}{%
    \begin{tabular}{@{}lcccc|cccc@{}}
        \toprule
         & \multicolumn{4}{c}{val} & \multicolumn{4}{c}{test} \\ 
        \cmidrule(lr){2-5} \cmidrule(lr){6-9}
        Backbone & R@1 & R@3 & R@5 & Ave & R@1 & R@3 & R@5 & Ave \\ \midrule
         CLIP~\cite{radford2021learning}      &62.7 & 86.6 & 92.8 & 80.7 & 71.3& 92.4 & 96.2 & 86.6 \\ 
       BLIP~\cite{li2022blip}       &\textbf{63.1} & \textbf{87.3} & \textbf{93.4} & \textbf{81.3} & \textbf{73.1}& \textbf{93.6} & \textbf{96.9} & \textbf{87.9} \\ 
        \bottomrule
    \end{tabular}%
    }
\end{table}
